# Adaptive Development of Koncepts in Virtual Animats: Insights into the Development of Knowledge

Carlos Gershenson
School of Cognitive and Computer Sciences
University of Sussex
C.Gershenson@sussex.ac.uk

*Abstract*

*As a part of our effort for studying the evolution and development of cognition, we present results derived from synthetic experimentations in a virtual laboratory where animats develop koncepts adaptively and ground their meaning through action. We introduce the term "koncept" to avoid confusions and ambiguity derived from the wide use of the word "concept". We present the models which our animats use for abstracting koncepts from perceptions, plastically adapt koncepts, and associate koncepts with actions. On a more philosophical vein, we suggest that knowledge is a property of a cognitive system, not an element, and therefore observer-dependent.*

## 1. Introduction

We are interested in the evolution, development, and acquisition of cognition (Gershenson, 2001). Let us first see possible answers to the question "How knowledge is acquired?". We can see that we need to perceive what we know, so we could say first that knowledge is acquired from perception. But we should note, first that perception in natural systems is *adaptive* (Kohler, 1962), and second that it is adaptive through action (Held, 1965). If we are interested in studying the acquisition, development, and evolution of knowledge, we need to begin studying, modelling, and understanding the *adaptive* processes which lead to successful and efficient perception and action for increasing the probabilities of survival of a system according to its ecological niche. An artificial cognitive system should *develop* in order to *adapt* with the same versatility as a living organism to its specific environment.

It has been proposed that meanings are perceptually grounded (Jackendoff, 1983; Gärdenfors, 2000) and learned by association (de Saussure, 1966; Gärdenfors, 2000, pp. 187-9). We could say that a cognitive system, natural or artificial, *acts* in order to give *meaning*[1] to its perceptions.

---

[1]Not only linguistic meaning, but also perceptual meaning. For example, we can say that a fly has a meaning *for* a frog if the frog eats flies. This meaning is related to the act of eating, and the internal need hunger. Of course, this meaning will be very different from the meaning a fly has for us.

Gärdenfors (2000, p. 47) writes: "Knowledge about categories is grounded in sensorimotor regions of the brain since damage to a particular sensorimotor region destroys the processes of categories that use this region in the analysis of perception." In concordance with this, the models for knowledge development we present here extract regularities from perceptions, and then ground their meaning through action. Such an approach dissolves the symbol grounding problem (Harnad, 1990), since we can identify symbols grounded not only on other symbols, but on perception and action.

We will attempt to study knowledge in a synthetic way (Steels, 1995): building artificial systems to contrast our theories will give us understanding of the phenomena we are modelling, and also it will allow us to engineer artificial systems with the capabilities of the phenomena we are modelling.

In the present work we focus on the acquisition and development of "koncepts". In the next section, we present a brief review of some work related with ours. In Section 3 we introduce the term koncept to avoid confusions brought by the wide use of the word concept. In Section 4 we present the "Knowledge Emerging from Behaviour" Architecture (KEBA), and models it uses for abstracting koncepts from perceptions, for plasticity of koncepts, and for associating koncepts with actions. We then introduce a virtual laboratory, accessible via Internet, which we developed for contrasting KEBA in virtual animats. In Section 6 we discuss observations made from simulations carried out in our virtual laboratory. We finally describe our future work and conclude with some philosophical implications.

## 2. Related Work

There has been a great deal of work in automatic categorization (*e.g.* clustering algorithms, self-organized maps (Kohonen, 1995)), but with different purposes than the ones we have. Usually work in this area is related with classification and automatization of tasks, and not with the study of cognition. Our model does categorization of perceptual inputs, but attempting to illuminate how this takes place in natural cognitive systems.

Scheier and Lambrinos (1996) developed a robot which distinguishes between different categories of objects. This is achieved by linking haptic and visual systems. Their robot is able to distinguish between pucks with a cross on them and conductive material and without them. They use neural networks to process the information from a camera (visual) and conductance sensors (haptic), and integrate them. Unfortunately, neural networks deliver no clear idea of how concepts might be developed. This project has great value for adaptive behaviour, but we would like to obtain more information about the development of cognition, and the complexity of the architecture prevents us from this, since it was designed with a different purpose.

It is a similar case with the work of Nolfi and Tani (1999). They developed a model and implemented it in software agents for extracting regularities and making predictions using neural networks. Again, this work is valuable in its area, but since it was not intended for studying the development of cognition, the models deliver not much more information than a natural cognitive system.

Agents with minimally cognitive behaviour (Beer, 1996; Slocum, Downey, and Beer, 2000) have been artificially evolved in order to study with the aid of dynamical systems theory simple artificial cognitive systems delivering very interesting results. The work of Seth (1998) is also in this spirit: he evolves the activation functions of direct links between sensors and motors in Braitenberg-style software agents for action selection tasks. These works explore the question of the evolution of cognition. We are also interested in the development and acquisition of it, but their findings and ideas are very much related to our goals.

All the works previously mentioned do not speak about concepts. They deal successfully with cognition at a lower level. They study more direct links between perception and action. Concepts can be identified when the relationships between perception and action are more complex. These works and ours have different approaches to slightly different things, but they are all useful to get a wider image of the development and evolution of cognition.

The work carried out at the ARTI Lab in Brussels (Steels 1996; Steels and Vogt, 1997) has produced robots which successfully ground meanings through discrimination games. But they consider that concepts have already been acquired, so they are linking these predefined concepts with meanings. We are also interested also in the acquisition and development of concepts.

Gärdenfors (2000) has developed a very interesting and broad theory of concepts based on geometry. But this theory has not been implemented in an artificial system yet, in order to be adjusted and contrasted in a synthetic way.

## 3. Koncepts

We can identify two types of categorization (Gärdenfors, 2000): continuous and discrete. The first one depends on sensation and perception, and the second one on concepts. Concepts can be seen as a *discretization* of the perceptual space.

Since concepts have been studied from very different perspectives (*e.g.* Peacocke, 1992; Gärdenfors, 2000), people might have different understandings of what do others mean by 'concept'. In order to avoid confusions, we will use the word 'koncept' to denote what we are studying.

Koncepts are abstractions (discretizations) of perceptions or of other koncepts. Of course, they are not limited to humans or to animals. Attempting to give an operational notion of koncept, we could say that if a system can "detect" different instantiations of an object, it has a koncept of that object, since it is able to abstract different perceptual states with one koncept. But, this can only be contrasted if: a) the system uses natural language to tell us "yes, I have a koncept of that object"[2], or b) the system performs a specific behaviour or action whenever *we* believe that the system *uses* a koncept. As all operational notions, this is observer-dependent.

Koncepts are emergent properties of cognitive systems, *i.e.* we <perceive|define> them. This is, you cannot "open" the system and search for the koncepts, because you will find nothing. They emerge from the interactions of the components of the system and the interactions of the system with its environment. Koncepts are in the observer, not in the system (unless if the system is the observer (self-contemplation)). It is because of this that we do not

---

[2]Suppose we are Martians and we have just arrived to earth: How would we judge if humans have koncepts without the use of language?

expect to find any neurophysiological correlation with any konceptual model. In other words, natural cognitive systems have koncepts not as objects, but as properties. At a chemical or neuronal level we do not expect to find any straight correlation with koncepts. These should be studied at a different level (Gärdenfors, 2000), as emerging from elements of lower levels (Gershenson, 2002)

As an example, we could say that frogs (Lettvin *et. al.*, 1959) have koncepts of small, dark, fast-moving objects which they identify with food, since in their ecological niche flies are the only objects which satisfy this condition. But they do not have a koncept of 'fly', first of all because they do not speak English, and second because there are objects which are not flies and activate this koncept. We can only *judge* that frogs *have* koncepts through their behaviour: if they snap their tongue when they perceive small, dark, fast-moving objects, we can say that they have a koncept of it, since they do not snap their tongue in presence of other objects. We can conclude that regularities in perception leading to regularities in action are a *necessary* condition for having koncepts. The *sufficient* conditions are subjective, since it depends on the observer to judge if a cognitive system *has* koncepts or not. This will be only if the observer can *identify* them emerging, independently of the implementation of the cognitive system.

It is useful to have a language to identify koncepts, but this sometimes carries confusions, and we should always be aware while studying koncepts in nonlinguistic cognitive systems we are the ones who are naming the koncepts. If the cognitive systems have no linguistic abilities, how could they name koncepts?

Another issue is that our language itself brings ambiguity (Wittgenstein, 1918) into deciding if a system uses knowledge or not. For example, in a certain context, we can say that humans are born knowing how to breathe. But in another context we could argue that only humans posses knowledge. Then fish are not born knowing how to swim? But would we agree in saying that some trees know when spring has come? That a soda can machine knows when a coin has been inserted? It is left to the observer to decide... And also we should be aware that our conception of knowledge will be determined by the context in which we are speaking (Gershenson, in preparation).

## 4. KEBA

We had previously developed the recursive model for koncept abstraction to be presented shortly. We improved it and we also developed the adaptive linking of koncepts to actions. We have named our models as KEBA, which stands for "Knowledge Emerging from Behaviour" Architecture.

We do not suggest that natural cognitive systems develop their knowledge in the same way as our models do, we are just intending to explore the principles of this development in cognitive systems in general. We are interested in *observing* the same properties that natural cognitive systems *exhibit*.

### 4.1. Abstracting Koncepts from Perception

We developed a model which generates recursively regularities from sensory input. These sensations include perceptual, introceptual (sensing the internal medium (*e.g.* hunger, thirst)), and propioceptual (sensing motor actions and body). We add noise to these signals attempting to approach inexactitudes in real environments (Jakobi, Husbands, and Harvey, 1995). The information from the sensors is mapped to the interval [0..1] and each input is taken as a koncept (protokoncept) of the zero level. Each one determines a coordinate of a multidimensional vector, as defined in multidimensional logic (Gershenson, 1998; 1999), which will be considered as the coordinates of the centres of new koncepts. These koncepts are generated in dependance of the experiences of the animat, combining sensory inputs, and are of level 1. They can create the coordinates of multidimensional vectors of koncepts of level 2, which can determine coordinates at a third level, and so on. The coordinates determine the centres of open balls (Gershenson, 1999), which are fuzzy neighbourhoods that along with two radiuses $r_1$ and $r_2$ determine the linear fuzzy membership function (1) of the distance d (calculated with Pythagoras) of a point to the centre of the open ball.

$$C(d) = \begin{cases} 1 & if\ d \leq r_1 \\ 0 & if\ d \geq r_2 \\ 1 - \dfrac{d - r_1}{r_2 - r_1} & if\ r_1 < d < r_2 \end{cases} \qquad (1)$$

Each koncept in KEBA has a value v, an activation a, and a stability s. The values of the koncepts at level zero are the signals arriving from the sensors. The values of the koncepts of levels n>0 are determined by how far the activations of the koncepts of level n-1 are from the centre of the koncept, namely C(d) (1). The activation represents the history of v's and is determined by (2):

$$a = \frac{v + A^n \iota}{1 + A^n \iota} \qquad (2)$$

where A is the activation potential raised to the power of the level n, and $\iota$ gives an importance to the previous experiences, acting as a persistence factor. With $A^n$, the activation of a koncept will take more importance from previous values if the koncept is in a higher level, and therefore will have a slower decay.

The koncepts active at time t in the level n determine a multidimensional vector, which might fall into an open ball of a koncept of the level n+1, altering its value as in (1), but if not creates a new koncept at level n+1 with centre at the coordinates determined by the activations and value of one. If the activations of the koncepts at level n do not fall into the open ball of a koncept of the level n+1, the latter is updated with v=0.

The stability of the koncepts represents the change of the activation. The value of s is bounded to the interval [0..1]. It is determined by (3)

$$s = s + \kappa - |a(t) - a(t-1)| \qquad (3)$$

where κ is the speed of the stability. It does not require to be higher than the noise. If κ is small, koncepts will be abstracted only when the environment is more stable. The larger the value of κ, the more details of the environment will be abstracted, reaching a maximum at κ=1.

The values of v, a, and s are updated recursively for the level n+1, creating new koncepts when necessary, only when all the koncepts of the level n are stable (s=1) and more than one koncept of the level n is active (a>0). Figure 1 shows an example of a koncept configuration of KEBA.

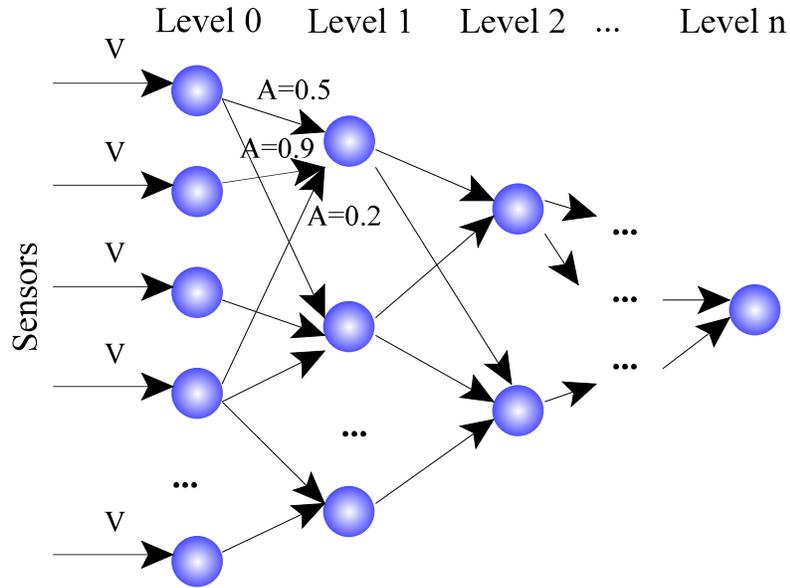

Figure 1. Example. Activations at level n determine the values of koncepts at level n+1

## 4.2. Plasticity of Koncepts

Each time step the vectors representing the centres of the koncepts are adjusted accordingly to (4):

$$\bar{x}(t+1)=\bar{x}(t)+\eta v(t)(\bar{a}(t)-\bar{x}(t)) \qquad (4)$$

where $\underline{x}$ is the centre of a koncept, $\underline{a}$ is the vector determined by the activations of the parents of the koncept, v is its actual value, and η determines the size of the adjustment.

If an actual value is inside a koncept, and the distance d to the centre of the koncepts is greater than the noise, then radiuses of the koncept are adapted as shown in (5):

$$r_1 = r_1 - \zeta \ \ if \ d \leq \frac{r_1}{2}$$

$$r_2 = r_2 - \zeta \ \ if \ \frac{r_1}{2} < d \leq r_1$$

$$r_1 = r_1 + \zeta \ \ if \ r_1 < d \leq \frac{r_1 + r_2}{2} \qquad (5)$$

$$r_2 = r_2 + \zeta \ \ if \ \frac{r_1 + r_2}{2} < d \leq r_2$$

where $\zeta$ is the amount of the adjustment.

This plasticity of koncepts allows slight adjustments for slight changes in the environment. For example, if the property of an object (*e.g.* a colour) represented as a dimension of the koncept changes slowly, the koncept will refer to the same object when it will have a drastically different property (colour), instead of having a different koncept.

### 4.3. Association of Koncepts with Actions

If we want to observe the properties KEBA, the developed koncepts should be *used*. One way of using them is by linking them to actions. We do this by a simple reinforcement learning scheme.

Each koncept has a link associated to a potential action. These are randomly initiated with medium values for protokoncepts, and for koncepts of levels higher than zero they are averaged between the links of their parents and a random medium value. If the greatest value of the links corresponds to the link of the actual action, and there has been a positive stimulus (satisfaction), then the link is incremented. This also occurs if the greatest link is different from the actual action and there has been a negative stimulus. The link is decremented otherwise: if the greatest link corresponds to the actual action and there has been a negative stimulus, or if there has been a positive stimulus when the greatest link and the actual action differ.

At this moment, since we are testing our models with simple actions (*e.g.* eat, drink), the increments and decrements are a simple step function (the link will be one if there is an increment, zero if there is a decrement). For more complex actions, hyperbolic convergence (Gershenson and González, 2000) could be used.

For deciding an action, every koncept with $v > 0$ contributes to a sum according to (6):

$$act_i += v(act_{ci})(lev+1)^2 \qquad (6)$$

where $act_i$ is the total activation for action I, $act_{ci}$ is the strength of the link (similar to a weight in an artificial neural network), and lev is the level in which the koncept is. Therefore, koncepts at higher levels will contribute more to the action selection.

After the results from all the koncepts have been summed, the action will be just the maximum of all $act_i$'s.

Once koncepts are *used* to select actions (but they are not as effective as a specially designed action selection mechanism), we can say that they have acquired a *meaning* for the artificial system. And we could identify meaning of not only consummatory behaviours, but also of appetitive (*e.g.* approach food), since koncepts which would be related with appetitive meanings would still be active when the associations of koncepts with actions takes place through a positive or negative stimulus. But also because of this, the appetitive meanings would be acquired slower than consummatory ones.

### 4.4. Properties of KEBA

KEBA recursively generates data structures (koncepts) which represent regularities in perceptions and previously generated regularities. The koncepts of KEBA are metaphors of koncepts we observe in natural cognitive systems. They plastically adapt to slight changes, and they provide links to actions, which we argue provides them with certain meaning. Koncepts in KEBA *represent* not only objects, but also, at a lower level, properties, and at a higher level, events, situations, and contexts. We argue that in cognitive systems, natural and artificial, the distinction of properties, objects, situations, etc., is made at a semantic/linguistic level. But at a konceptual level, they have the same essential properties, only that some are more general, others are more particular.

We also argue that KEBA is suitable for studying inductions[3]. What we would call a proposition or a sentence (*e.g.* "if A then B"), can be generated as koncepts of higher levels, which are determined by other koncepts which we would identify with objects or events (*e.g.* "A", "B").

KEBA is still in development, but we believe that it provides a promising perspective for studying and understanding the acquisition, development, and evolution of cognition. At this stage, we have linked subconceptual (*e.g.* signals from receptors) and conceptual levels (Gärdenfors, 2000). The link between the conceptual and semantic level is seen as future work. This would provide non-symbolic semantic computational mechanisms, in the sense that the meaning is not predefined.

## 5. A Virtual Laboratory for Contrasting KEBA

We have been developing a virtual laboratory (following the ideas proposed in Gershenson, González and Negrete (2000)) for contrasting our models with the aid of virtual animats. It can be downloaded (including source code in Java and Java3D) from http://www.cogs.sussex.ac.uk/users/carlos/keb.

The laboratory consists of a toroidal virtual environment, where different phenomena (at this stage animats, food sources, rocks, lightnings and rains) can be created and deleted deterministically or randomly, and which have a determined behaviour. Screenshots of the virtual environment can be seen in Figure 3 and Figure 4. Each phenomenon (including animats) has associated qualitative values ([0..1]) of redness, greenness, blueness, odour+ (positive, pleasant), odour- (negative, unpleasant), loudness, flavour+, flavour-, and hardness.

---

[3]We believe that inductions are made not only by humans, but also by other natural cognitive systems.

Animats through their sensors can perceive only from the environment these values, with random noise[4] (controlled by the user). The first six can be perceived when the objects are at a distance smaller than a radius of perception, and the last three when the animat is touching the phenomenon.

The animats have three internal variables: energy, hunger, and thirst; which take values from the interval [0..1]. Hunger and thirst are increased linearly through time, and are reduced when an appropriate action is executed in the presence of an appropriate stimulus (*e.g.* eat while touching food, drink while touching rain), and are increased in the stimulus is inappropriate for the action (*e.g.* eat while touching a rock). When the values of hunger or thirst are high, the energy value decreases. When they are low, the energy is increased. When the energy value reaches zero, the animat dies. Animats can perform the following actions: eat, drink, and no action. Their locomotion pattern can be switched manually to one of the following behaviours: wander, circular, and static. The maximum levels in KEBA can be assigned to each animat.

Animats receive as protokoncepts signals from sensors, the internal variables, and the current actions. An example of how koncepts are determined can be seen in Figure 2. For clarity, we reduce here the number of dimensions to the colour components, but actually the koncepts of the zero level (protokoncepts) will generate koncepts with fifteen dimensions. For this example, we can say that when an animat is perceiving only rain, the sensors generate a koncept in the first level, determined by the activations of the sensors (protokoncepts of zero level). And when an animat is perceiving only food, KEBA will generate a koncept in a fuzzy neighbourhood determined by the activations of the protokoncepts. But if in our virtual environment rains turn into foods, the activations of the koncepts rain (already decayed when the food has appeared, but still active), and food, will determine a koncept of level two, which *we* could interpret as "if food then rain". It depends on the animat if this knowledge will be of some *use*. But we consider that the way koncepts are recursively abstracted, provides an elegant way of developing *inferences* (Gärdenfors, 2002, ch. 6).

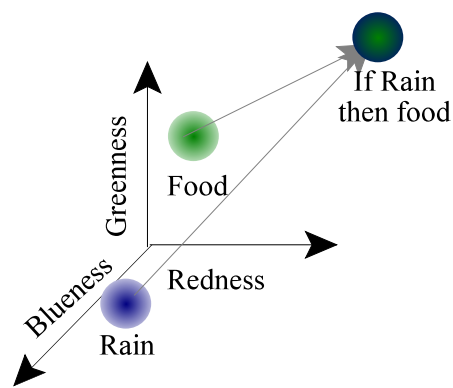

Figure 2. Example of koncepts in animats.

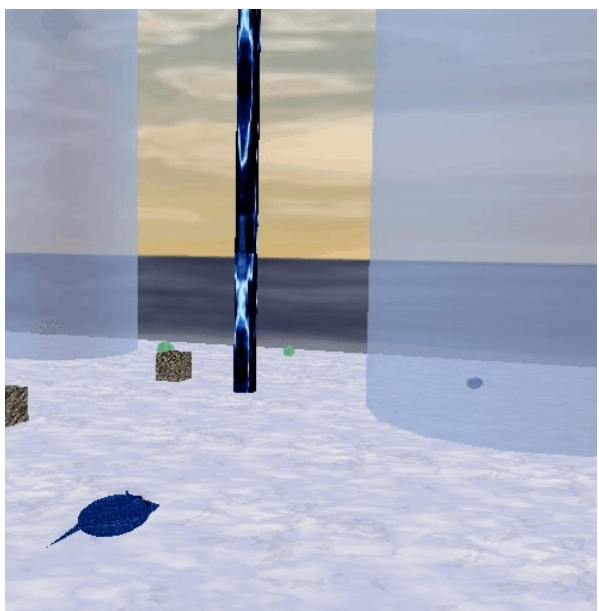 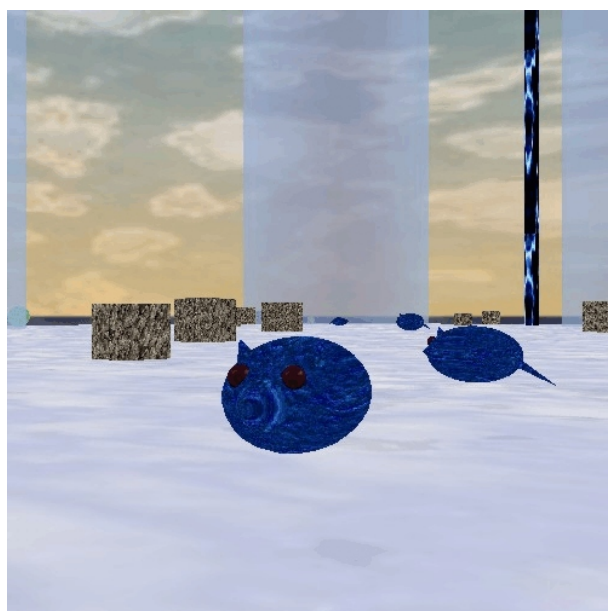

Figure 3. Screenshot of virtual environment.   Figure 4. Screenshot of virtual environment.

Lightnings turn into rains after 10 time cycles, and rains turn into foods after 50 time cycles. Foods decrease their size when they are eaten. Animats may feed from other animats, reducing their energy value.

Simulations can be saved and loaded from files simplifying the experimentation in our virtual laboratory.

## 6. Simulations

In this section we present observations taken from some simulations carried out in our virtual laboratory. The reader is encouraged to download the software via Internet and test the observations presented here.

In Figure 5 we can appreciate the dynamics for the internal state 'hunger' in a typical run for animats using KEBA, with random action selection, and with no actions. They were set with a wandering behaviour in an environment generating stimuli randomly. The animat performing no actions has a linear increment of hunger. For the other animats, their hunger is increased when they try to eat an inappropriate stimulus (*e.g.* a rock), and reduced when they eat next to a food source or another animat. The animat with random action selection will have a varying value of hunger, since he may try to eat appropriate and inappropriate stimuli, but also not to eat them. The animat using KEBA, learns from his past experience, which by reinforcement links koncepts to actions, and therefore in the long run will have a better performance than the other animats. He learns to "eat the right thing", avoiding to eat inappropriate stimuli. We describe this as the grounding of the meaning of the koncepts which determine the successful feeding of the animat. Animats without actions always die after about 4500 sim cycles, and animats with random action selection usually die earlier. Animats using KEBA might also die, they are not perfect, since they are not designed for a specific task, but their life expectancy is much higher. An animat with an evolved Braitenberg-style control mechanism or with an evolved neural network controller surely would make accurate links between the sensors and actuators for a specific environment, but we are not interested directly in the survival of the animats, but in their robust cognitive development. A combination of both probably would yield interesting results. But we have seen that animats using KEBA do *adapt* in randomly generated environments, in the sense that they learn to do the appropriate actions in the appropriate circumstances.

Even when this task might seem simple, it is not. It is easy for us to identify objects we perceive ("hey, look, a tree"). But it is not easy at all to build a system which can respond in the same way than we do to its perceptions. Also it is not well understood how do *we* do this (otherwise we would just imitate it). We believe that KEBA might contribute in illuminating how abstractions in natural and artificial cognitive systems can take place.

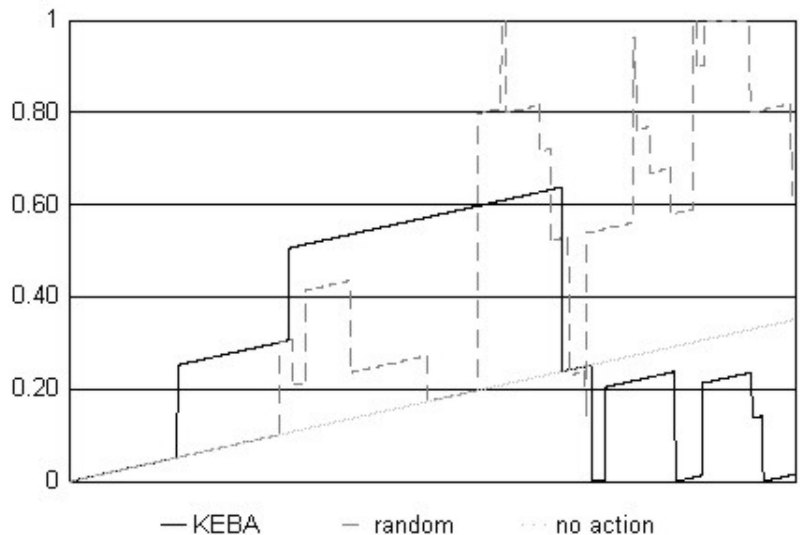

Figure 5. Hunger for a typical run of animats using KEBA, with random action selection, and with no actions.

We observed the development of koncepts in KEBA varying the amount of noise. Without noise, koncepts should correspond directly to external phenomena, so there are less koncepts generated. As we increment noise, the amount of koncepts is increased because of the noise. But when the noise is high, the amount of koncepts is reduced again, because they are less stable, and it is not possible to generate koncepts of higher levels. Animats survive even for noise values up to 0.5, since for surviving there is no need of koncepts of higher levels. but when the noise is too high, they are not able to abstract koncepts, so they cannot learn from their past experience. Their performance could be compared to an animat with random action selection.

Animats with different number of maximum abstraction levels also are able to survive in their environment, but there is a difference depending on the maximum levels allowed in KEBA. The issue is that the koncepts of the higher levels would be useful for actions more complicated than the ones we have implemented yet. In other words, an animat with only proto-koncepts would be able to survive, but would not be able to develop knowledge. Animats need a complex body (*e.g.* a real robot) to exploit koncepts in complex actions and behaviours.

Our animats use one of three possible behaviour patterns for movement: wander, circle, and static. In the same order, the complexity of the perceived environment is decreased, since the changes in the Umwelt[5] of the animat make its sensations to be more diverse, and thus its koncepts to be more varied. As the Umwelt becomes more complex and diverse, the koncepts are less well-shaped. A possibility for "shaping" these koncepts would be to share them between animats through some form of *communication*, and this would provide certain uniformity to the Umwelts of different animats in the same environment.

We have also observed that different animats have similar koncepts at lower levels but these tend to differ as the levels are increased. This is mainly because the koncepts at higher levels depend more on the specific history of the animat, while the ones in lower levels depend more on the environment of the animat. Again, communication seems to be a possible way for sharing and using the abstract koncepts located in higher levels.

## 7. Future Work

We are still developing KEBA, and there are many issues we should attend. One of them is the "forgetting" of koncepts. Because of hardware limitations, the memory of a computer becomes full if too many koncepts are abstracted. If we are running a simulation with several animats, this occurs even faster. One way of overcoming this would be to "forget" koncepts which have not been used, and were created more as an episode than as an abstraction. This would also illuminate more how the koncepts are created in KEBA.

We are also interested in the development of the logic of a particular environment (for example, in our virtual laboratory, IF lightning THEN rain, IF rain THEN food, IF lightning THEN food), and how cognitive systems might exploit this knowledge. For this, we need to devise more complicated actions, in order to be able to *observe* the possible advantages of such knowledge.

Another issue we are interesting in studying is the sharing of koncepts. This should be by some form of communication, but this communication should occur through the same sensors. This might provide new insights into the origins of communication, and its necessary conditions.

In the long term, KEBA might also have impact in the study of evolution and development of language, since we believe that koncepts are a necessary condition for *using* a language. This would be interesting also for exploring the following question: how similar are the konceptual structures of different individuals using the same language?

In general, KEBA is an aid for exploring the necessary conditions for the development and evolution of cognition.

---

[5]Term introduced by von Uexkull (1985) to denote the particular environments of animals, from *their* perspective.

## 8. Conclusions

We propose that koncepts in natural cognitive systems are <generated|emerging> from regularities in perception and action. In natural cognitive systems koncepts are not physical structures, but emergent properties. We have developed artificial animats which are able to generate data structures which we identify as koncepts, from regularities in perception and previous koncepts using our "Knowledge Emerging from Behaviour" Architecture (KEBA). We argue that the *meaning* of these koncepts is grounded through action. Also, KEBA supports the idea that cognition, natural or artificial, <u>needs</u> to be adaptive (*e.g.* Maturana and Varela, 1987).

We believe that animals and humans *develop* koncepts in a similar way as our animats, but there is no direct way of testing this (we cannot chop a brain and see koncepts), but we need to do it as observers, unavoidingly carrying certain subjectivity. Of course, the actual mechanisms involved in the development of koncepts may be different as the ones we propose, but we are obtaining an idea of the necessary conditions required for this development to take place. Imagine that the actual mechanism in natural brains can be represented with a function in $\mathbb{R}^n$. There are an infinitude of functions in $\mathbb{R}^{n+1}$ which exactly follow that function (as there are an infinitude of planes ($\mathbb{R}^2$) which cross any straight line ($\mathbb{R}^1$)). It is very hard to say if our approximation to a function is "the real one", but for sure we can obtain a general idea of it with different approximations. Our proposal is just one of these approximations. And we believe that it is useful for studying the necessary conditions for the development of koncepts, meanings, and cognition in general.

We believe that we should study not the *origins* of cognition (*e.g.* Donald, 1991), but its <development|evolution>. This is because we cannot draw a sharp border between non-cognitive and cognitive systems. We can just identify gradually different levels and complexities of cognition.

We suggest that knowledge is a *property* of a cognitive system, not an element. Therefore, it is exhibited, more than possessed. This has consequences in the judgement considering which systems "have" cognition. If we take the stance of cognition as a possession, <u>we</u> feel that we posses knowledge because we can communicate such feeling. But even if other systems can have the same feeling, we cannot know this. If take the stance of cognition as a property, we should observe the behaviour of a system (human, animal, robot, society), and judge its cognition according to the properties we observe, not to its constitutive elements. This depends on how strict, or how complex, our definition of cognition is that more or less systems will be considered as cognitive, and at which degree.

Being inspired by our experiments with KEBA, we say that *we* do not "perceive" <u>the</u> phenomena in our environment. We do not see directly *a* cup. Under appropriate light conditions, photons bounce on the surface of objects, and these photons activate receptors in our retina. This leads to the activation of a circuit which we relate with the word "cup", "taza", "чашка" or whatever. In a similar way, animats with KEBA receive their sensor signals, which then activate koncepts. We generate the koncepts, and we name them. Therefore koncepts are dependent of their user.

We should also note the importance of *experience* in the development of knowledge. A cognitive system needs to experience its environment in order to develop certain knowledge of it. Therefore, we can see that knowledge is tied to the environment of the cognitive system. This makes knowledge to be *relative* and *context-dependent* (Gershenson, in preparation).